\documentclass[conference]{IEEEtran}
\usepackage[utf8]{inputenc}
\IEEEoverridecommandlockouts
\usepackage{cite}
\usepackage{amsmath,amssymb,amsfonts}
\usepackage{algorithmic}
\usepackage{graphicx}
\usepackage{textcomp}
\usepackage{xcolor}
\usepackage{dblfloatfix}
\usepackage{tabularx}
\usepackage{capt-of}
\usepackage{booktabs}
\usepackage[resetlabels]{multibib}
\usepackage{placeins}
\usepackage{seqsplit}
\newcites{app}{Test Collection Papers}
\newcolumntype{Y}{>{\raggedright\arraybackslash}X}
\def\BibTeX{{\rm B\kern-.05em{\sc i\kern-.025em b}\kern-.08em
    T\kern-.1667em\lower.7ex\hbox{E}\kern-.125emX}}
\begin{document}

\title{AquiLLM: Evaluating Faithfulness in Open-Weight RAG-LLM Systems for Scientific Research
}

\author{
\IEEEauthorblockN{
Bernie Boscoe\IEEEauthorrefmark{1},
Srinath Saikrishnan\IEEEauthorrefmark{4},
Vikram Seenivasan\IEEEauthorrefmark{2},
Jack Stark\IEEEauthorrefmark{2},
Andrew Lizarraga\IEEEauthorrefmark{3}, \\
Morgan Himes\IEEEauthorrefmark{2},
Jonathan Soriano\IEEEauthorrefmark{2},
PJ Allen\IEEEauthorrefmark{1},
Tuan Do\IEEEauthorrefmark{2}
}

\IEEEauthorblockA{
\IEEEauthorrefmark{1}Dept. of Computer Science, Southern Oregon University, USA\\
\IEEEauthorrefmark{2}Dept. of Physics \& Astronomy, University of California, Los Angeles, USA\\
\IEEEauthorrefmark{3}Dept. of Statistics, University of California, Los Angeles, USA\\
\IEEEauthorrefmark{4}Dept. of Computer Science, University of California, Los Angeles, USA
}

\IEEEauthorblockA{\small
boscoeb@sou.edu, srinathsai22@ucla.edu, vikrams25@ucla.edu,\\
jstark@astro.ucla.edu, andrewlizarraga@ucla.edu, jsoriano@astro.ucla.edu, \\
 morganhimes@ucla.edu, pricep@sou.edu, tdo@astro.ucla.edu
}
}    
\maketitle

\begin{abstract}
Scientific research increasingly relies on large, heterogeneous data sources, motivating interest in retrieval-augmented generation (RAG) systems that provide natural language access to scientific knowledge and research workflows.
Researchers are exploring the viability of these systems as natural language interfaces for document search and for generating analysis code and pipeline components.
At the same time, concerns about data privacy and control over research infrastructure have motivated interest in open-weight models and open-source deployments hosted within research institutions.

In astronomy, this development follows a long history of computational infrastructure development, from archival databases and Structured Query Language (SQL)-based systems to large language model (LLM)-assisted research tools. This paper presents a domain-expert evaluation of faithfulness for AquiLLM, an open-weight, offline RAG-LLM platform designed to support scientific research groups in the use and preservation of tacit and formal knowledge. 

We define faithfulness as the extent to which generated responses remain grounded in retrieved scientific context without unsupported claims or omissions. We report results from an astronomy case study evaluating AquiLLM across retrieval and scientific analysis tasks. AquiLLM performs most reliably on explicit retrieval-oriented questions grounded in the RAG collection, while faithfulness degrades for queries requiring synthesis or ambiguity resolution. These results highlight both the promise and limitations of open-weight RAG-LLM systems for scientific research and demonstrate the importance of domain-expert evaluation beyond standard benchmark leaderboards.

\end{abstract}


\section{Introduction}
As scientific data systems grow in scale and complexity, natural language interfaces can lower barriers for researchers seeking to access increasingly heterogeneous forms of knowledge \cite{androutsopoulos1995}. Modern scientific workflows require navigating data and code distributed across large, often non-interoperable platforms. Research knowledge also exists in less formal, tacit forms, including unpublished manuscripts, meeting notes, emails, software documentation, and Jupyter notebooks. The fragmentation of these sources creates substantial cognitive and technical overhead for researchers.

Astronomy has long been a leader in developing computational data infrastructure and large-scale scientific archives in response to the growing scale and complexity of observational data. Projects such as the Sloan Digital Sky Survey transformed access to astronomical data through searchable repositories, database-driven architectures, and SQL-based query interfaces developed through collaborations working at the intersection of astronomy and computer science \cite{szalay2000a}. These systems helped establish a broader paradigm for data-intensive science in which large observational datasets could be explored remotely through web portals and structured database queries. Although such SQL-based systems significantly democratized access to scientific data, they still require familiarity with database schemas and query syntax. 

Natural language interfaces offer a way to reduce this complexity by allowing researchers to interact with scientific systems conversationally without requiring specialized query languages or software expertise \cite{zaouiseghroucheni2025, collins2013}. Retrieval-augmented large language model (RAG-LLM) systems are particularly relevant in this transition from structured SQL query systems toward natural-language scientific interfaces \cite{hey2009}. Recent advances in LLMs, RAGs, and accessible computational infrastructure have accelerated interest across scientific domains \cite{skarlinski2024, lewis2021}. Despite this momentum, many research groups remain cautious about relying on commercial AI platforms for workflows involving unpublished data and proprietary analysis. Concerns surrounding privacy, reproducibility, and cost have motivated growing interest in open-weight models and locally deployed RAG-LLM systems.

AquiLLM was developed as an open-source open-weight platform designed to help research groups collectively preserve and interact with both formal and tacit knowledge \cite{boscoe2026, campbell2025}. We use \textit{open-weight} to refer to models whose trained parameters are publicly available, while the underlying training data, code, and methodology remain unavailable. These models can still be locally deployed and fine-tuned. The AquiLLM system allows researchers to create personal or group collections and query them using natural language to support information retrieval and onboarding of new group members. Additionally, AquiLLM is modular, allowing research groups to adapt models to domain-specific workflows and retrieval over specialized research corpora. 

Open-weight language models are proliferating rapidly, yet they are still evaluated through benchmark leaderboards and general-purpose language tasks \cite{liang2023}. 
While these benchmarks provide useful comparative metrics, performance on standardized evaluations does not necessarily reflect reliability in scientific contexts. Scientific applications require systems to accurately reflect source material and avoid unsupported inference. In this work, we define \textit{faithfulness} as the extent to which generated responses remain grounded in retrieved scientific context without fabrication or unsupported inference. Although human evaluation remains the gold standard for assessing faithfulness, such evaluations are difficult and costly to conduct at scale \cite{malin2025}.

We conduct a domain-expert evaluation of faithfulness in which astronomers assess AquiLLM responses across categories of scientific queries created from their research group’s own data and documentation. The study includes questions involving both factual retrieval and comparative scientific analysis. Through this evaluation, we examine how faithfulness varies across different forms of scientific interaction and identify limitations that are not captured by standard benchmark evaluations. This work establishes a baseline for understanding the limitations of open-weight offline RAG-LLM systems in domain science and highlights the importance of domain-expert evaluation for improving trustworthy RAG-LLM systems.

This paper makes three contributions. First, we present AquiLLM, a deployed, modular, open-weight RAG-LLM system (Section~\ref{sec:system}) that research groups can adopt directly for local, privacy-preserving access to their own scientific knowledge. Second, we introduce a domain-expert evaluation methodology built around five query categories spanning varying levels of retrieval complexity and reasoning (Section~\ref{sec:querydesign}), which other research groups can reuse to evaluate faithfulness in their own scientific RAG deployments. Third, we report empirical findings from applying this methodology to an astronomy research group, identifying six recurring failure modes and showing that faithfulness degrades specifically for cross-source synthesis and comparative reasoning rather than uniformly across query types (Section~\ref{sec:results}). Because this is an initial, single-group, single-domain study, we treat these contributions as a baseline evaluation rather than a general benchmark; Section~\ref{sec:limitations} details the resulting scope constraints, including the absence of a controlled comparison against an alternative RAG system.

\section{Related Work}
 Early work on natural language interfaces to databases explored how users could query conversationally rather than through formal query languages such as SQL \cite{androutsopoulos1995}. Natural language processing (NLP) evolved alongside the growth of cyberinfrastructure platforms designed to support data-intensive research workflows \cite{gray2004}. Research software engineers, whether identified by this title or otherwise, have played a central role in the building and maintenance of these systems to aid computational research \cite{szalay2001}. Before the research software engineer (RSE) role became more formalized, domain scientists often relied on graduate students, collaborations with computer scientists, or researchers working across both disciplinary and computational domains to develop scientific software and infrastructure. 

Contemporary LLM interfaces emerged from decades of scientific cyberinfrastructure development, extending earlier database systems into conversational knowledge interfaces \cite{luo2025}. Recent RAG architectures combine LLMs with retrieval systems \cite{ersoy2025}, using vector databases and semantic search methods to ground responses in domain-specific document collections. In scientific settings, these systems have been explored for literature search, scientific question answering, and interaction with technical documentation and code \cite{guu2020, skarlinski2024, nguyen2023}.
Many domain-specific RAG systems focus on grounded generation to reduce hallucinations and improve the reliability of generated responses \cite{ji2023}. AquiLLM builds on these approaches while emphasizing the preservation of tacit research knowledge and research-group context \cite{polanyi1967}.

Much of the recent development in LLM interfaces has occurred through commercial AI platforms and hosted services. In contrast, open-weight models and locally developed RAG-LLM systems give research groups direct control over how their scientific knowledge systems are configured, audited, and maintained, without depending on an external provider's infrastructure or terms of service. Benchmarking in other scientific domains has similarly found that open-source models can match closed-source performance while offering greater transparency, reproducibility, cost-effectiveness, and data privacy \cite{yang2025}. AquiLLM builds on this perspective through a modular, locally-controlled, open-source approach to a RAG-LLM system.

Central to these systems is response reliability. LLMs may generate plausible but fabricated or misleading information \cite{shuster2021}. Prior work has explored concepts including correctness, factuality, faithfulness, and groundedness for evaluating generated responses \cite{akbar2024, khan2026}. In scientific contexts, these concerns are important because research workflows depend on accurate interpretation of source material and trustworthy communication of technical knowledge.

Scientific applications require higher standards of reliability than general-purpose conversational systems. Plausible but unsupported claims or incorrect synthesis across documents undermines trust in LLM-enabled workflows. Retrieval-augmented systems aim to reduce these risks by grounding responses in retrieved evidence, though questions remain regarding how faithfully such systems represent scientific source material in practice \cite{maynez2020, tamber2025}.
Benchmark leaderboards provide useful comparative metrics but do not adequately capture reliability in domain-specific scientific settings \cite{liang2023}. Evaluating specialized scientific knowledge requires contextual understanding and familiarity with research workflows that automated metrics may fail to detect \cite{ji2023}.
\begin{figure*}[!t]
  \centering
  \includegraphics[width=\textwidth]{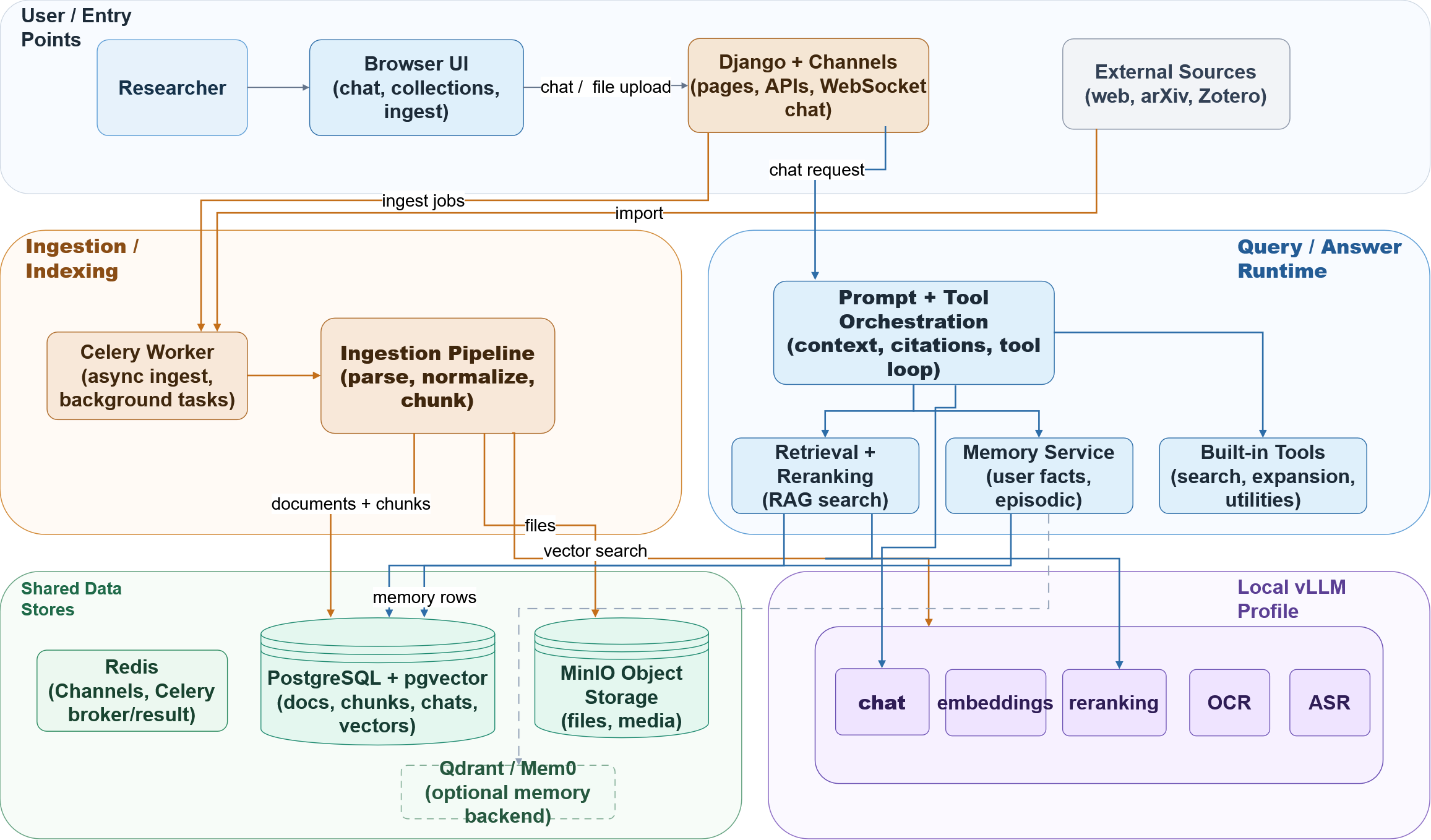}
  \caption{High-level AquiLLM architecture. Background workers handle document ingestion and indexing, while runtime requests pass through a prompt orchestration layer coordinating retrieval, memory, and tool use. Local inference is served through vLLM using separate models for chat, embedding, reranking, and multimodal processing.}
  \label{fig:aquillm-arch}
\end{figure*}
As a result, human evaluation remains an important approach for assessing faithfulness and reliability in scientific systems. However, expert annotation is difficult and costly. These challenges motivate contextual evaluation frameworks in which domain experts assess responses grounded in their own research-group knowledge collections, rather than relying solely on standardized benchmark tasks developed by model builders.
\section{AquiLLM System Overview}
\label{sec:system}
AquiLLM is an open-source browser-based RAG-LLM tool designed to support research groups through natural language interaction with scientific knowledge collections \cite{boscoe2026}. Users can create personal or shared collections containing both published and private data, then query these materials conversationally. The system supports information retrieval while preserving tacit knowledge embedded in scientific workflows and research collaboration. Originally developed for astronomy and environmental science research groups, AquiLLM is designed as a general framework for research-group knowledge systems.

AquiLLM source code and deployment configuration are openly available through the GitHub project repository, allowing research groups to build and deploy their own AquiLLM instances. Persistent public deployments remain constrained by the computational resources required to support locally hosted open-weight models and multimodal retrieval workloads. Our current research deployments operate on dedicated GPU infrastructure to support interactive workflows and concurrent users. We periodically provide live demonstrations while exploring more computationally accessible deployment configurations for research groups.

\subsection{Astronomy Case Study}
For this study, we evaluated an astronomy-specific deployment of AquiLLM built around astronomy and machine learning research collections. Participating astronomers prepared a dedicated collection consisting of survey documentation and publications used in their research.
For this initial study, we focused on formal scientific artifacts, including unpublished manuscripts and technical documents, rather than broader tacit workflow and collaborative research materials. This allowed us to establish a baseline evaluation of faithfulness before incorporating more complex research-group knowledge collections.

\subsection{System Architecture}
AquiLLM uses a modular RAG architecture organized around ingestion, retrieval, orchestration, and inference components. Documents are uploaded through a browser-based interface, processed asynchronously, chunked, and indexed for semantic retrieval, shown in Fig.~\ref{fig:aquillm-arch}. The system supports heterogeneous scientific materials including PDFs, figures, and LaTeX-based mathematical notation. During query-time execution, retrieved document chunks are combined with user prompts through an orchestration layer that grounds responses in domain-specific collections.

The architecture supports local deployment using open-weight models rather than commercial AI APIs. AquiLLM separates generation, embedding, reranking, and OCR into independent components, allowing retrieval and multimodal workloads to operate separately from the primary conversational model. This modular design improves extensibility across research domains while allowing research groups to modify retrieval pipelines, models, and document collections without changing the overall framework. The astronomy deployment evaluated in this study was hosted on Jetstream2 using dedicated H100 GPU resources. Recent development has focused on local embeddings, reranking, vLLM-based inference, and memory-layer support for multimodal retrieval and private deployment workflows.

\subsection{Model Selection and Configuration}
The astronomy deployment evaluated in this study used locally hosted Qwen-based open-weight models integrated through the AquiLLM inference pipeline. The primary conversational model was \texttt{\seqsplit{Qwen3.5-27B-Claude-4.6-OS-Auto-Variable-Thinking}} \cite{davidau2026}, an independent community fine-tune of the official Qwen3.5-27B base model \cite{qwen3report2025}, selected for stronger reasoning behavior and more consistent tool usage than the base model in our own informal testing. This fine-tune was produced by a third-party contributor (not Alibaba, the Qwen family's publisher) by further training the base model on a "Claude-4.6-OS" dataset of Claude-derived examples, intended to introduce Claude-style variable-length reasoning ("thinking") behavior into the base model. Its behavior is therefore not directly covered by Alibaba's own published Qwen benchmarks \cite{qwen3report2025}, and, to our knowledge, no independent technical report or benchmark evaluation of this specific fine-tune exists; our selection was based on informal comparison against the base Qwen3.5-27B model on our own workflows rather than a published evaluation. Both the base model and this fine-tune are released under a permissive Apache 2.0 license and support local deployment and further fine-tuning.

The deployment additionally used separate Qwen-based models for embedding, reranking, and multimodal processing, allowing retrieval and document-processing workloads to operate independently from the conversational model. This modular approach improved flexibility while supporting heterogeneous scientific workflows and multimodal retrieval tasks.

Model selection was guided by reasoning quality, retrieval performance, computational cost, and feasibility within locally managed infrastructure. The selected configuration fit within the available H100 GPU resources on Jetstream2 while supporting retrieval-augmented prompting and concurrent interactive usage. Larger models were considered, but many exceeded practical GPU memory limits or introduced latency that reduced usability. The deployment therefore represents a balance between model capability and deployability within institutional computing environments. Future work will explore domain-specific fine-tuning using astronomy corpora and evaluation feedback collected through AquiLLM.

\subsection{Retrieval and Indexing Configuration}
\label{sec:retrievalconfig}
To support reproducibility, Table~\ref{tab:ragconfig} reports the retrieval, chunking, and reranking configuration used during the evaluation deployment. Documents are split into overlapping character-based chunks, embedded, and indexed for retrieval. At query time, AquiLLM performs a hybrid search that pools candidates from dense vector similarity and lexical (trigram) matching, expands the candidate pool by a fixed multiplier beyond the requested top-$k$, and reranks the pooled candidates before passing the top results to the generation model. These values correspond to the project's default configuration at the time of the study (source-controlled configuration as of March 31, 2026, immediately preceding the evaluation period) and are exposed as environment variables so that individual deployments can retune them for their own corpora \cite{boscoe2026}.

\begin{table}[t]
\centering
\caption{Retrieval, chunking, and reranking configuration used during the evaluation deployment.}
\label{tab:ragconfig}
\begin{tabular}{ll}
\hline
\textbf{Setting} & \textbf{Value} \\
\hline
Embedding model & Qwen3-VL-Embedding-2B (1024-dim) \\
Reranker model & Qwen3-VL-Reranker-2B \\
Chunk size & 2048 characters \\
Chunk overlap & 384 characters \\
Vector retrieval top-$k$ & 12 \\
Lexical (trigram) retrieval top-$k$ & 12 \\
Candidate fan-out multiplier & 3$\times$ \\
\hline
\end{tabular}
\end{table}

Generated responses are additionally subject to citation enforcement: the generation pipeline requires cited references to map only to chunks actually returned by the retrieval tools, with a fail-safe fallback that produces a cited response if the model does not otherwise comply. In the AquiLLM interface, citations in generated responses are clickable and open a slide-out panel showing the originating document with the cited passage highlighted, allowing users to directly verify a claim against its source material. This citation-enforcement mechanism was implemented as of March 31, 2026, the same source-controlled snapshot reported in Table~\ref{tab:ragconfig}, and was therefore active throughout the evaluation.

\section{Study Design and Methods}
\subsection{Participants}
The study involved five participants from the same astronomy research group with expertise in observational astronomy and machine learning workflows. Participants regularly collaborated on publications, datasets, and research software associated with the group’s astronomy projects and had prior familiarity with AquiLLM for approximately one year.

\subsection{Corpus and Collection Construction}
\label{sec:corpus}
Participants constructed a dedicated AquiLLM collection consisting of 31 scientific documents and technical references associated with the group’s astronomy and machine learning workflows. The collection included survey documentation and instrumentation papers related to the Hyper Suprime-Cam Survey (HSC), the Euclid Mission, and the Rubin Observatory Legacy Survey of Space and Time (LSST), along with the group's own publications, unpublished work, and technical references connected to the group’s GalaxiesML dataset research efforts. 
The 31-document collection comprised 15 peer-reviewed journal articles, 4 conference papers, and 12 additional technical resources (preprints, mission and survey documentation, and dataset descriptions), listed in full in the accompanying test-collection bibliography. We did not separately measure the corpus's indexed storage footprint (e.g., chunk or token count); Section~\ref{sec:scalability} discusses how faithfulness and retrieval behavior may change for substantially larger collections, which this study does not evaluate.

The corpus combined straightforward factual materials with more complex research-oriented content. Evaluation tasks included both direct retrieval questions and more challenging interpretation and comparative analysis tasks involving redshift prediction models, emission-line reconstruction, and machine learning performance metrics. The collection was uploaded into AquiLLM for shared use by participants.
\subsection{Evaluation Procedure}
\label{sec:evalprocedure}
Participants contributed an initial pool of candidate domain-specific questions spanning the major projects in their research group (survey and instrumentation documentation, the group's GalaxiesML dataset and MMAE model work, and general machine-learning/statistics onboarding topics), together with draft reference answers grounded in their own research activities, datasets, workflows, and publications. From this pool, the research team selected ten representative questions (Table~\ref{tab:question_metadata} in the Appendix) for the primary evaluation. Questions were selected to keep the rating workload manageable for a five-participant panel while still spanning all five query categories (Section~\ref{sec:querydesign}) and each of the group's major project areas, rather than being sampled at random or exhaustively evaluating the full candidate pool. Participants then collaboratively established consensus reference answers for these ten questions; reference answers were discussed and finalized as a group during the research group's regular astronomy research meetings, rather than by a single arbiter.

To account for variability in generation behavior, each participant independently submitted each of the ten questions to AquiLLM up to three times through their own chat session, generating and then rating their own responses through the built-in evaluation form; no two participants rated the same generated response text. This yielded up to 150 independent generation-and-rating instances (10 questions $\times$ 5 participants $\times$ 3 submissions); after removing incomplete submissions, the final evaluation dataset contained 141 rated responses (Table~\ref{tab:querycats}), consistent with the per-question counts of 12--15 completed ratings reported in Section~\ref{sec:agreement}. Participants reviewed generated responses and assessed faithfulness relative to the retrieved scientific context and their domain expertise. Responses were evaluated using a five-point scale measuring factual accuracy, grounding, preservation of scientific context, and avoidance of unsupported claims. Participants also provided qualitative written feedback describing strengths, ambiguities, omissions, or inaccuracies.

Participants interacted with AquiLLM through the browser-based interface used during normal workflows. Ratings and qualitative comments were submitted through a built-in evaluation form, anonymized using participant identifiers, and exported for analysis. AquiLLM had no access to the public internet via external tool calls during this study. This reflects the privacy-sensitive scientific environments in which the system may be deployed, where external web access may be undesirable or prohibited. Restricting internet access also ensured that generated responses were grounded exclusively in the curated test collection used for evaluation, rather than external web content or SEO-optimized sources of varying reliability.

\subsection{Query Design}
\label{sec:querydesign}
Evaluation questions were grouped into five query categories: factual retrieval, metadata lookup, onboarding and conceptual explanation, scientific interpretation, and comparative scientific analysis. These categories captured varying levels of retrieval complexity, synthesis, and domain reasoning within astronomy workflows.

Factual retrieval questions involved direct lookup of scientific information from retrieved documents. Metadata lookup questions focused on survey structure, observational channels, release information, and dataset characteristics. Onboarding and conceptual explanation questions evaluated whether the system could explain terminology or concepts useful for newcomers. Scientific interpretation questions required understanding scientific outputs and model behavior, while comparative scientific analysis questions involved comparing models, datasets, or performance metrics across contexts.

The final evaluation dataset consisted of 141 annotated responses distributed across the five query categories shown in Table~\ref{tab:querycats}.
\begin{table}[t]
\centering
\caption{Distribution of query categories in the evaluation dataset.}
\label{tab:querycats}
\begin{tabular}{lr}
\hline
Query Category & Count \\
\hline
Factual Retrieval & 60 \\
Comparative Scientific Analysis & 39 \\
Scientific Interpretation & 15 \\
Metadata Lookup & 14 \\
Onboarding / Conceptual Explanation & 13 \\
\hline
\end{tabular}
\end{table}

\subsection{Evaluation Criteria}
\label{sec:criteria}
In this study, faithfulness was defined as the extent to which generated responses remained grounded in scientific context without fabrication or unsupported inference. Participants evaluated AquiLLM responses against collaboratively established reference answers for the selected evaluation questions.

Evaluation emphasized consistency with reference answers and an absence of hallucinated claims. Responses were evaluated using a five-point Likert-style scale ranging from low to high faithfulness. The evaluation form presented this scale as a plain 1--5 rating control without predefined verbal anchors for each point (e.g., no fixed text such as ``fully faithful'' or ``not faithful'' attached to a specific value); interpretation of the scale instead relied on participants' shared, group-level understanding of faithfulness developed through approximately one year of prior collaborative use of AquiLLM. Each rating was accompanied by required qualitative written feedback describing strengths, ambiguities, omissions, or inaccuracies, which we used to recover the specific reasoning behind low or high scores (Section~\ref{sec:failuremodes}). We treat the absence of formal verbal anchors as a limitation of the current evaluation instrument (Section~\ref{sec:limitations}). Participants generated and rated their own responses independently through their own AquiLLM chat session (Section~\ref{sec:evalprocedure}): the evaluation interface did not display other participants' ratings or generated responses, so raters could not see or be influenced by one another's scores, and no two participants ever rated the same generated response text. Individual ratings were not reconciled into a single consensus value, unlike the reference answers, which were discussed as a group; the dataset retains each participant's rating separately, which is what enables the descriptive consistency analysis in Section~\ref{sec:agreement}, though that analysis cannot separate rater subjectivity from genuine response-to-response variation for the reason just described. Because evaluators were members of the same research group and collaboratively established reference answers, the study emphasized qualitative agreement and contextual scientific interpretation rather than formal inter-rater reliability metrics; Section~\ref{sec:agreement} nonetheless reports a descriptive analysis of rating consistency across participants for each question, computed from the existing multi-rater data.

\section{Results}
\label{sec:results}
Across 141 cleaned responses (140 non-missing ratings), AquiLLM demonstrated generally strong faithfulness performance, with a mean rating of 3.80 (median = 4.0, SD = 1.24). Most responses were evaluated positively, with 68.6\% receiving ratings of 4--5, while 17.9\% received ratings of 1--2.

The system performed most reliably on retrieval-oriented questions grounded in single scientific documents, while performance became less stable for synthesis and comparative analysis tasks requiring integration across multiple sources. Qualitative comments described AquiLLM as useful and scientifically helpful, though participants also identified omissions and unsupported claims in more complex query settings.
\subsection{Performance Across Query Types}
As shown in Table~\ref{tab:query_type_performance}, query-type variation was the strongest signal in the evaluation. The most robust contrast, drawn from the two categories built from multiple underlying questions, is between Factual Retrieval (4 questions, $n{=}60$, mean 4.17) and Comparative Scientific Analysis (3 questions, $n{=}39$, mean 2.97), indicating that cross-source synthesis and comparison remain the primary failure surface. The remaining three categories, Scientific Interpretation, Metadata Lookup, and Onboarding/Conceptual Explanation, are each represented by a single underlying question in this study (Table~\ref{tab:question_metadata}), so their category-level means are single-question results rather than generalizable category effects. In particular, the high mean for Scientific Interpretation (4.73) reflects performance on one specific question about MMAE emission-line reconstruction and should not be read as evidence that reasoning-oriented questions generally outperform retrieval-oriented ones; the abstract's summary claim that AquiLLM performs most reliably on retrieval-oriented questions rests on the multi-question Factual Retrieval/Comparative Scientific Analysis contrast, not on this single-question result. Onboarding and metadata questions were generally helpful but showed more moderate reliability, with errors more often related to precision and scope than outright hallucination.

\begin{table*}[t]
\centering
\small
\caption{Performance across the five query types. Factual Retrieval and Comparative Scientific Analysis aggregate multiple questions each (4 and 3, respectively); the remaining three categories are each a single question (Table~\ref{tab:question_metadata}) repeated across participants and submissions, and their means should be read accordingly.}
\label{tab:query_type_performance}
\setlength{\tabcolsep}{4pt}
\begin{tabularx}{\textwidth}{l c c Y Y}
\toprule
\textbf{Query Type} & \textbf{n} & \textbf{Mean} & \textbf{Observed Strengths} & \textbf{Common Failure Modes} \\
\midrule
Factual Retrieval (4 questions) & 60 & 4.17 & Accurate grounded responses; strong direct evidence retrieval; generally high consistency & Occasional omission; missing qualifiers/details \\
Comparative Scientific Analysis (3 questions) & 39 & 2.97 & Attempts cross-result comparison and tradeoff framing & Inconsistent grounding across sources; selective omission; incomplete synthesis \\
Scientific Interpretation (1 question) & 15 & 4.73 & Helpful synthesis and interpretation when context is clear; strong explanatory coherence & Unsupported inference in edge cases; occasional overreach beyond cited evidence \\
Metadata Lookup (1 question) & 14 & 3.64 & Useful schema/channel/parameter lookup support; good orientation to dataset structure & Partial precision; assumptions not always explicit \\
Onboarding / Conceptual Explanation (1 question) & 13 & 3.62 & Useful onboarding support for foundational concepts; accessible explanations & Overgeneralization; simplified but under-specified answers \\
\bottomrule
\end{tabularx}
\end{table*}

\subsection{Rating Consistency Across Independent Attempts}
\label{sec:agreement}
Because each participant independently generated and rated their own responses rather than rating a shared set of fixed responses (Section~\ref{sec:evalprocedure}), no two ratings in this dataset correspond to the same generated response text. As a result, the spread of ratings for a given question reflects a mixture of two sources that cannot be separated with this data: genuine variation in how faithfully AquiLLM answered that question across repeated, independent attempts, and differences in how strictly individual participants rated. This second source is itself an expected one: although human evaluation is generally considered the gold standard for assessing faithfulness, it is also known to be prone to subjective variation between evaluators \cite{malin2025}. We therefore report the following as a descriptive measure of overall rating consistency, computed directly from the released evaluation dataset, rather than a formal inter-rater reliability statistic, which would require multiple raters to score the same fixed response (Section~\ref{sec:criteria}).

Across the ten evaluation questions (12--15 completed ratings per question, from up to five participants each), the mean within-question standard deviation of ratings was 0.90 (range 0.35--1.46 across questions), and 77.8\% of individual ratings fell within one point of that question's mean rating.

Consistency was highest for the three highest-scoring questions (Q6--Q8 in Table~\ref{tab:question_metadata}), which per their own classification span Scientific Interpretation, Factual Retrieval, and a mixed retrieval/reasoning task rather than being uniformly retrieval-oriented, where 100\% of ratings fell within one point of the question mean, and lowest for two Comparative Scientific Analysis questions (Q4 and Q9), where only 42.9\% and 46.2\% of ratings, respectively, fell within one point of the mean. Because response variation and rater variation are completely confounded in this dataset, as described above, we cannot attribute this pattern to either source specifically: it may reflect AquiLLM's own output being less stable for comparative-analysis questions from one independent attempt to the next, participants disagreeing more about how to score genuinely ambiguous synthesis attempts, or some mixture of both, and this analysis provides no way to distinguish between these possibilities. We note only that lower consistency co-occurs with the same category, Comparative Scientific Analysis, where mean faithfulness was also lowest (Section~\ref{sec:results}), without claiming a specific causal explanation for that co-occurrence. We report this analysis as an exploratory substitute for a formal reliability metric, not a replacement for one (Section~\ref{sec:limitations}).

\subsection{Failure Modes}
\label{sec:failuremodes}
To better understand why some responses received low faithfulness ratings, we categorized evaluator-identified errors into six recurring failure modes.

\subsubsection{Hallucinated Scientific Details}
The model occasionally introduced unsupported numeric or technical claims while sounding scientifically plausible. In one comparison between two redshift prediction models, the system reported a 2.4x bias gap when the source-supported value was 1.2x.

\subsubsection{Unsupported Synthesis}
When evidence was fragmented across documents, the model sometimes inferred conclusions without sufficient grounding. In one case involving \texttt{c\_model\_mag} computation, evaluators reported that the system initially stated it could not locate the method, but then generated a profile-fitting explanation not supported by the retrieved materials.

\subsubsection{Retrieval Mismatch}
Some failures reflected incorrect retrieval alignment or incomplete anchoring to the requested source context. Evaluators flagged responses that referenced unspecified tables or documents, or stated that information could not be found even though it was present in the supplied collection.

\subsubsection{Subtle Factual Drift}
Not all failures were overt hallucinations; several responses were nearly correct but contained shifted values, missing qualifiers, or scope drift. One evaluator noted that a response correctly identified the scientific trend under discussion but reported the wrong scatter value and omitted important conditional details.

\subsubsection{Overconfident Interpretation}
The model occasionally presented uncertain or weakly supported claims with excessive confidence. In conceptual explanation tasks, such as discussions of z-band naming conventions, evaluators observed assertive interpretations that were not strongly supported by the retrieved scientific context.

\subsubsection{Failure Under Ambiguity}
For underspecified prompts, the system did not always request clarification before answering. Evaluators noted that some survey-channel questions were answered for a specific astronomy survey without first confirming which survey the user intended.

Overall, the dominant risk observed in the evaluation was not random error, but credible-sounding overreach, particularly in synthesis and comparative analysis tasks requiring integration across multiple sources. This pattern aligns with the lower average faithfulness ratings observed for Comparative Scientific Analysis questions.

\subsection{Expert Perspectives}
Participants found AquiLLM most useful for direct retrieval-oriented tasks involving numeric values, survey metadata, or information grounded within a single document. Several described the system as effective for quickly locating scientific details and navigating technical materials.

Trust decreased for tasks requiring reasoning across multiple papers or fragmented scientific context. Participants observed greater variability and lower reliability for synthesis-oriented questions, often producing plausible but only partially grounded responses. One participant summarized this behavior as, ``when asked to reason across papers it fell apart.''

Participants also emphasized the importance of query specificity. Broad or underspecified prompts often caused retrieval of excessive or weakly related information spanning multiple astronomy surveys rather than the intended dataset.

Another recurring issue was variability across repeated queries. Repeated submissions of the same question occasionally produced inconsistent retrieval behavior or conflicting answers. These inconsistencies raised questions about retrieval persistence, memory effects, and tool-use stability.

\subsection{Sources of Failure}
\label{sec:sourcesoffailure}
Observed failures reflected interactions between retrieval quality, orchestration behavior, model reasoning, and implementation limitations. Some errors resulted from incomplete retrieval or insufficient ranking of relevant scientific context, particularly for synthesis-oriented questions spanning multiple documents.

Other inconsistencies appeared related to tool-calling and orchestration behavior. Participants observed variability in whether retrieval tools were invoked across repeated generations of the same query, including cases where responses were generated without visible retrieval after earlier successful retrieval attempts.

Several issues also appeared to reflect implementation-level limitations rather than failures of the underlying language model itself, including inconsistent retrieval visibility, repeated-query variability, and poor handling of ambiguous prompts.

Overall, the evaluation suggests that improving faithfulness in systems such as AquiLLM will require advances in retrieval quality, tool-use consistency, ambiguity handling, and system-level robustness rather than foundation models alone.
\section{Discussion}
\subsection{What Faithfulness Means in Scientific Contexts}

The results of this study suggest that faithfulness in offline scientific RAG-LLM systems differs substantially from general-purpose notions of chatbot helpfulness or conversational fluency. In scientific workflows, responses must remain grounded in source material and avoid unsupported inference that could affect interpretation or reproducibility. As a result, scientific trust depends not only on whether a response is correct, but also on whether researchers can understand how and why the response was generated from the underlying evidence. This becomes particularly important in scientific domains where measurements and analysis methods differ across research groups investigating similar phenomena. One long-term goal of AquiLLM is to support this kind of contextual scientific reasoning by distinguishing between competing methodological interpretations and explicitly communicating these differences within generated responses.

Participants in the study frequently evaluated responses in terms of traceability and provenance rather than stylistic quality alone. Responses that accurately reflected retrieved documents and consistent with known references were generally viewed as trustworthy, even when incomplete. In contrast, plausible-sounding synthesis without clear grounding often reduced confidence, particularly in comparative analysis tasks involving multiple papers or datasets. Evaluators also emphasized the importance of uncertainty awareness, noting that overly confident explanations could become misleading when evidence was ambiguous or only partially retrieved.

These findings suggest that scientific RAG-LLM systems require stronger support for transparent retrieval behavior, source attribution, ambiguity handling, and calibrated uncertainty, than is typically emphasized in consumer conversational AI systems. More broadly, the evaluation highlights that scientific faithfulness is not solely a property of the language model itself, but emerges from the interactions between retrieval systems, orchestration pipelines, domain context, and user expectations surrounding evidence and reproducibility.

\subsection{Implications for Scientific Cyberinfrastructure}
Systems such as AquiLLM can be viewed as part of a longer evolution in scientific cyberinfrastructure, particularly within astronomy. Earlier platforms such as the Sloan Digital Sky Survey (SDSS), CasJobs, and web-based archive systems expanded scientific access through searchable databases and SQL-driven interfaces. RAG-LLM systems represent a potential new interface layer by enabling conversational interaction with publications, datasets, technical documentation, and collaborative research knowledge.

However, the results of this study suggest that important reliability constraints remain unresolved. While AquiLLM performed relatively well for retrieval-oriented tasks, synthesis-heavy and comparative reasoning questions exposed limitations in grounding, retrieval consistency, and tool orchestration. These findings suggest that future scientific NLP-based infrastructure will require advances not only in foundation models, but also in retrieval pipelines, provenance tracking, and transparent reasoning workflows.

\subsection{Scalability and Generalization Potential}
\label{sec:scalability}
This study evaluated AquiLLM against a modest, curated collection of 31 documents; we have not yet systematically measured how retrieval quality, latency, or faithfulness change as collections grow toward the scale of active research-group archives (hundreds to thousands of documents) or shared institutional repositories (potentially hundreds of thousands). AquiLLM's retrieval layer combines dense vector search with lexical matching over indexed chunks (Section~\ref{sec:retrievalconfig}), and its modular separation of ingestion, embedding, reranking, and generation (Fig.~\ref{fig:aquillm-arch}) is intended to let each component be scaled or replaced independently as corpus size grows. These are architectural design goals rather than measured outcomes, however, and larger or noisier collections may increase retrieval ambiguity in ways that compound the comparative-analysis failure modes already observed at small scale (Section~\ref{sec:failuremodes}); we did not evaluate this directly.

Generalization beyond this single astronomy research group raises a related but distinct question. AquiLLM's modular design lets each research group configure its own collection, retrieval settings, and (within hardware constraints) underlying models, and we expect the query-category evaluation methodology used in this study (Section~\ref{sec:querydesign}) to transfer directly to other domains. Faithfulness thresholds, common failure modes, and appropriate evaluation panels are nonetheless likely to be domain-specific: a collection of numerically dense survey papers, as used here, differs in structure and ambiguity from collections dominated by qualitative methods, code, or heterogeneous file types. We therefore treat this study as establishing a reusable evaluation methodology rather than a result that generalizes numerically to other domains without re-evaluation; Section~\ref{sec:futurework} outlines planned evaluation across additional research groups and domains to test this directly.

\subsection{Open-Weight LLMs for Research Communities}
The study additionally highlights the potential importance of open-weight RAG-LLM systems for research communities. Locally deployable systems provide greater institutional control over data, infrastructure, reproducibility, and long-term preservation compared to externally managed commercial AI platforms. Although frontier commercial systems may outperform smaller open-weight models, research groups may still prefer open systems that can be modified, and preserved within research computing environments.

\subsection{Lessons for RSE Teams}
For research software engineers evaluating or deploying similar offline RAG-LLM systems, three practical lessons emerged from this study. First, retrieval-oriented and synthesis-oriented queries behave differently enough that they warrant separate evaluation and, likely, separate design attention; a single aggregate faithfulness score can mask large differences between query types (Table~\ref{tab:query_type_performance}). Second, requiring a short free-text justification alongside each numeric rating, even without a formalized rubric or predefined scale anchors, recovered most of the diagnostic value in this study (Section~\ref{sec:failuremodes}) and is a low-cost addition to any in-house evaluation form. Third, several of the observed failures traced back to retrieval and orchestration behavior rather than to the underlying language model (Section~\ref{sec:sourcesoffailure}), suggesting that RSE teams building similar systems should budget evaluation and debugging effort for the retrieval/orchestration layer specifically, not only for model selection.

\subsection{Human Expertise Remains Essential}
The evaluation demonstrates that domain experts remain essential for assessing scientific faithfulness in non-deterministic systems. Participants identified subtle inaccuracies that would likely remain invisible in benchmark evaluations. These findings suggest that benchmark performance alone is insufficient and expert-centered evaluation should remain central to future scientific RAG-LLM research.

\section{Limitations}
\label{sec:limitations}
This study has several limitations. First, the evaluation involved five astronomers from a single research group, which may limit broader generalization. Second, the study focused on a single scientific domain and a specific astronomy deployment of AquiLLM; other disciplines may exhibit different retrieval practices, terminology, and expectations surrounding scientific faithfulness (Section~\ref{sec:scalability}). Third, the evaluation examined a single open-weight model configuration within the AquiLLM architecture. Because the open-weight ecosystem is evolving rapidly, newer models and tools may produce substantially different behaviors and failure characteristics.
Although preliminary internal testing with commercially hosted models connected to AquiLLM suggested similar grounding and synthesis limitations, these configurations were not systematically evaluated in the present study.

Fourth, this study does not include a controlled comparison against an alternative RAG system or a component-level ablation isolating the contribution of reranking, retrieval fan-out, or orchestration. A direct comparison against a commercially hosted RAG system is complicated by the same privacy motivation that underlies AquiLLM's design: the evaluation corpus includes the research group's unpublished manuscripts and internal technical references (Section~\ref{sec:corpus}), and transmitting this material to an external commercial API would undermine the offline, privacy-preserving deployment model that this paper argues for. Restricting such a comparison to only the corpus's publicly available documents would avoid this problem but would no longer test the same corpus used throughout this study, weakening rather than strengthening the comparison. We therefore view two next steps as more coherent than a commercial-system comparison: (1) a component-level ablation conducted entirely within AquiLLM's own offline deployment, varying reranking and cross-document orchestration behavior while holding the retrieval and generation models fixed, targeted at the Comparative Scientific Analysis category where faithfulness failures and rating inconsistency both concentrated (Sections~\ref{sec:results} and~\ref{sec:agreement}); and (2) a comparison against another locally-deployable open-source RAG framework on the same corpus, which avoids the data-exposure problem of a commercial API. We plan to pursue both as follow-ups to this study.

Fifth, the five-point faithfulness scale used in the built-in evaluation form did not include predefined verbal anchors (Section~\ref{sec:criteria}). Moreover, because each participant generated and rated their own responses rather than rating a shared set of fixed responses (Section~\ref{sec:evalprocedure}), formal inter-rater reliability statistics are not computable from this dataset, not merely uncomputed: no two ratings correspond to the same response text, so any measured spread in ratings necessarily mixes rater subjectivity with genuine generation-to-generation variability that cannot be separated post hoc. The descriptive analysis in Section~\ref{sec:agreement} is intended as an initial, exploratory substitute, not a replacement for a validated rubric and a study design that would support a true reliability measure, such as multiple raters independently scoring the same fixed response.

\subsection*{Data and Code Availability}
AquiLLM's source code and deployment configuration are publicly available at \texttt{https://github.com/AquiLLM} \cite{boscoe2026}. We do not publicly release the evaluation dataset (questions, reference answers, ratings, and qualitative comments) at this time. This question set and its reference answers are intended to remain a held-out evaluation benchmark across the follow-up studies described in Section~\ref{sec:futurework}; publishing them openly now risks their incorporation into future model training corpora, which would contaminate the benchmark and undermine its usefulness for evaluating later open-weight models on the same questions. The dataset is available directly from the authors to individual researchers for reproducibility purposes, on the condition that it not be published or used, directly or indirectly, to train a model; we plan to release it publicly once the planned follow-up evaluation rounds are complete.

\subsection{Future Work}
\label{sec:futurework}
Our nearest-term priority is evaluation, not new system features. AquiLLM has since undergone a series of architectural updates, including local embedding and reranking, multimodal capabilities, and expanded memory support, described in a companion paper \cite{stark2026}. We plan to rerun this faithfulness study, using the same query categories and evaluation methodology (Section~\ref{sec:querydesign}), against the updated system to test directly whether these changes reduce the synthesis and comparative-analysis failure modes identified in Section~\ref{sec:failuremodes}. This rerun will specifically incorporate knowledge-graph-enhanced retrieval, extending AquiLLM's existing hybrid vector/lexical search (Section~\ref{sec:retrievalconfig}) with a graph of relationships between scientific concepts, entities, and their supporting evidence, since Comparative Scientific Analysis, the primary failure surface identified in Section~\ref{sec:results}, centers on cross-source synthesis and reasoning that a graph-structured representation is intended to support more directly than chunk-level retrieval alone. This rerun will be conducted alongside a broader component and configuration comparison, extending the ablation and open-source RAG comparison described in Section~\ref{sec:limitations} to measure how individual components, including retrieval, reranking, knowledge graphs, and memory, each contribute to performance, and we plan to extend the evaluation to additional research groups and domains, as well as to the tacit and informal knowledge artifacts (e.g., meeting notes, internal documentation) that \cite{stark2026} identifies as not yet evaluated for faithfulness.

Beyond this evaluation program, planned system development includes refining domain-specific knowledge-graph schemas and interactive graph exploration tools; extending provenance and citation tracing to graph relationships as well as individual passages; improving query interpretation, multi-part question handling, and ambiguity handling; more reliable tool orchestration and more robust multimodal document ingestion; continued latency and scalability work across retrieval, reranking, graph extraction, and local inference; domain-specific model fine-tuning; and methods for identifying when answers may be unreliable, ambiguous, or insufficiently supported by retrieved evidence. Ongoing work on prompt compression, additional multimodal embeddings, and longer-horizon memory planning is described in \cite{stark2026}.

\section{Conclusion}
Natural language interfaces are likely to become an important component of scientific cyberinfrastructure. Systems such as AquiLLM suggest that open-weight, offline retrieval-augmented language models can support scientific workflows through conversational interaction. In particular, retrieval-oriented tasks demonstrated promising levels of faithfulness and usability within the astronomy evaluation environment.

Additionally, the study demonstrates that scientific deployment of RAG-LLM systems requires rigorous evaluation within real research settings. Although AquiLLM performed well on retrieval-oriented tasks, the evaluation showed that maintaining faithfulness becomes substantially more difficult when systems must synthesize information across documents or operate under ambiguity. These findings highlight the importance of expert-centered evaluation for understanding the limitations of scientific RAG-LLM systems beyond standard benchmark performance.

\section*{Acknowledgment}
This work used Jetstream2~\cite{hancock2021} at Indiana University through allocation CIS260251 from the Advanced Cyberinfrastructure Coordination Ecosystem: Services \& Support (ACCESS) program~\cite{boerner2023}, which is supported by U.S. National Science Foundation grants \#2138259, \#2138286, \#2138307, \#2137603, and \#2138296. This research is also supported by National Artificial Intelligence Research Resource (NAIRR) Pilot allocations NAIRR240359 and NAIRR240464 and the Jetstream2 cloud resource supported by the National Science Foundation under Award NSF-OAC 2005506 at Indiana University.

This material is based upon work supported by the National Science Foundation under Award No.~2448094, the Alfred P. Sloan Foundation under Grant No.~G-2024-22720, and the UCLA DataX Institute.

\bibliographystyle{IEEEtran}
\bibliography{faithfulness26}

\newpage
\onecolumn
\appendices

\section{Deployment Cost and Hardware Requirements}
\label{sec:deploycost}
The astronomy deployment evaluated in this study ran on a single NVIDIA H100 GPU (80~GB), allocated through Jetstream2 via an ACCESS-CI allocation using the \texttt{g5.xl} instance flavor (20 vCPUs, 240~GB system RAM, 1$\times$ H100 GPU with 80~GB GPU memory), which Jetstream2 rates at 128 Service Units (SUs) per hour against the allocation. All inference components used in the study, the primary chat model, embedding, reranking, OCR, and transcription, were served as separate vLLM processes on this single GPU using memory-utilization partitioning (Section~\ref{sec:retrievalconfig}) to fit within its available memory, rather than requiring one GPU per component.

The SU-based rate above reflects usage against a grant-funded academic allocation rather than a direct cash payment by the research group. For research groups without access to an academic HPC allocation, Table~\ref{tab:deploycost} instead gives an approximate commercial-cloud equivalent cost for renting H100 GPU capacity, based on rates checked directly against provider pricing pages in September 2026\footnote{Specialist: RunPod pricing page (\texttt{runpod.io/pricing}), H100 PCIe/SXM on-demand, \$1.99--3.49/hr as of September 2026. Hyperscaler: Microsoft Azure's single-GPU \texttt{NC40ads\_H100\_v5} instance, \$6.98/hr on-demand (Azure pricing, September 2026); Amazon Web Services does not sell a standalone single-H100 instance, its \texttt{p5.48xlarge} instance bundles 8 H100 GPUs at \$55.04/hr on-demand (\$6.88/hr per GPU-equivalent), so the AWS figure is a computed per-GPU rate from an 8-GPU-minimum product rather than a directly purchasable single-GPU offering. Rates fluctuate and should be treated as indicative rather than precise.}. Specialist GPU cloud providers (e.g., RunPod) listed on-demand H100 rates around \$2--3.50 per GPU-hour; hyperscaler providers (Azure, AWS) listed single-H100 or per-GPU-equivalent rates around \$7--11 per GPU-hour. These configurations are not strictly comparable to each other or to the deployment evaluated in this study: Azure's single-GPU instance bundles roughly twice the vCPUs and system RAM of the \texttt{g5.xl} flavor used here, and the AWS figure requires committing to an entire 8-GPU node regardless of whether a research group needs that capacity. All figures also cover GPU compute only and exclude storage, networking, and engineering or maintenance time.

For a small research group (e.g., approximately seven users, comparable in size to the participant panel in this study plus its broader research group), usage is likely bursty rather than continuous: interactive queries occur intermittently, while document ingestion and indexing run asynchronously in the background. Table~\ref{tab:deploycost} therefore reports two usage scenarios: continuous (24/7) operation, which keeps the deployment immediately available and avoids model reload latency, and business-hours-only operation (approximately 8 hours per weekday), which reduces cost at the expense of availability outside that window.

\begin{table}[h]
\centering
\caption{Approximate commercial-cloud cost to run a single H100 GPU deployment, using on-demand rates checked directly against provider pricing pages in September 2026 (see text for provider-specific caveats).}
\label{tab:deploycost}
\begin{tabular}{lcc}
\hline
 & \textbf{Specialist cloud} & \textbf{Hyperscaler} \\
 & (\textasciitilde\$2--3.50/hr) & (\textasciitilde\$7--11/hr) \\
\hline
Continuous (24/7) & \$1,440--2,520/mo & \$5,040--7,920/mo \\
Business hours only\textsuperscript{*} & \$350--620/mo & \$1,230--1,940/mo \\
\hline
\end{tabular}

\vspace{2pt}
\footnotesize\textsuperscript{*}Approximately 8 hours/weekday, 22 weekdays/month ($\approx$176 hours).
\end{table}

These figures are indicative rather than precise: actual GPU pricing fluctuates, and the appropriate configuration depends on a research group's size, query volume, and availability requirements. They nonetheless suggest that a single-GPU, multi-component deployment of the kind evaluated in this study is within reach of a specialist-cloud budget of roughly several hundred to a few thousand dollars per month for a small research group, likely less expensive than the hyperscaler configurations in Table~\ref{tab:deploycost}, though the two are not directly comparable given the differences in bundled resources and minimum commitment described above. The deployment evaluated in this study itself was run without direct cloud-compute charges to the research group, against a grant-funded ACCESS-CI allocation instead; that allocation still represents a real, finite compute resource rather than a cost-free one.

\clearpage
\section{Evaluation Questions}
\centering
\begin{table*}[ht]

\caption{Metadata and classification of evaluation questions.}
\label{tab:question_metadata}

\footnotesize
\setlength{\tabcolsep}{3pt}
\renewcommand{\arraystretch}{1.15}

\begin{tabularx}{\textwidth}{
>{\raggedright\arraybackslash}X
>{\raggedright\arraybackslash}p{2.2cm}
>{\centering\arraybackslash}p{1.2cm}
>{\centering\arraybackslash}p{1.2cm}
>{\centering\arraybackslash}p{1.3cm}
>{\centering\arraybackslash}p{1.6cm}
}
\toprule

\textbf{Question} &
\textbf{Query Type} &
\textbf{Reasoning} &
\textbf{Synthesis} &
\textbf{Ambiguity} &
\textbf{Retrieval vs Reasoning} \\

\midrule

What are the channels of the galaxy data in this survey? &
Metadata Lookup &
Low &
No &
Yes &
Retrieval \\

Why is there a z-channel in photometric images and why is the redshift also denoted by z? &
Onboarding / Conceptual Explanation &
Medium &
Yes &
No &
Mixed \\

How are the c\_model\_mag magnitudes calculated? &
Factual Retrieval &
Medium &
Yes &
No &
Mixed \\

What are the spectroscopic survey changes from the second public data release to the third public data release of HSC? &
Comparative Scientific Analysis &
High &
Yes &
No &
Reasoning \\

How does the MMAE model's redshift prediction performance compare to CNN-based models in terms of scatter? &
Factual Retrieval &
Medium &
No &
No &
Retrieval \\

How do the shapes of emission lines reconstructed by the MMAE compare to those of the true lines? &
Scientific Interpretation &
High &
Yes &
No &
Reasoning \\

What is the maximum redshift of GalaxiesML-Spectra? &
Factual Retrieval &
Low &
No &
No &
Retrieval \\

At approximately what percentage of the GalaxiesML dataset do the LoRA model's performance metrics saturate? &
Factual Retrieval &
Medium &
Yes &
Yes &
Mixed \\

What is a drawback of using CNN-LoRA compared to CNN-Combo? &
Comparative Scientific Analysis &
High &
Yes &
No &
Reasoning \\

How does CNN-Base-Rev perform when compared to CNN-LoRA and CNN-LoRA-Rev on the Combo dataset? &
Comparative Scientific Analysis &
High &
Yes &
No &
Reasoning \\

\bottomrule
\end{tabularx}

\end{table*}
\FloatBarrier
\twocolumn

\nociteapp{*}
\bibliographystyleapp{IEEEtran}
\bibliographyapp{AquiLLM_Test_Collection}

\end{document}